\documentclass{bioinfo}
\copyrightyear{2005}
\pubyear{2005}

\usepackage{multirow}
\usepackage{amsmath}
\usepackage{amsfonts}
\usepackage{threeparttable}
\usepackage{url}
\usepackage{natbib}
\newcommand{\laga}{{\sc LaGE}}

\begin{document}
\firstpage{1}

\title[short Title]{\laga: A Java Framework to reconstruct Gene Regulatory Networks from Large-Scale Continues Expression Data}
\author[Sample \textit{et~al}]{Yang Lu\,$^{1}$, Mengying Wang\,$^{2}$,Kenny Q. Zhu\,$^{1}$ and Bo Yuan\,$^1$\footnote{to whom correspondence should be addressed}}
\address{$^{1}$Department of Computer Science, Shanghai Jiao Tong University, Shanghai, China\\
$^{2}$Software Engineering Institute, East China Normal University, Shanghai, China}

\history{Received on XXXXX; revised on XXXXX; accepted on XXXXX}

\editor{Associate Editor: XXXXXXX}

\maketitle

\begin{abstract}

\section{Summary:}

\laga~ is a systematic framework developed in Java. The motivation of \laga~ is to provide a scalable and parallel solution to reconstruct Gene Regulatory Networks (GRNs) from continuous gene expression data for very large amount of genes. The basic idea of our framework is motivated by the philosophy of divide-and-conquer\cite{LAMA}. Specifically, \laga~ recursively partitions genes into multiple overlapping communities with much smaller sizes, learns intra-community GRNs respectively before merge them altogether. Besides, the complete information of overlapping communities serves as the byproduct, which could be used to mine meaningful functional modules in biological networks.

\section{Availability:}

The source code and the supplementary documentation are available at http://202.120.33.37/LAGE/.

\section{Contact:} \href{luyang0415@sjtu.edu.cn}{luyang0415@sjtu.edu.cn}
\end{abstract}

\section{Introduction}

It is a key point in systems biology to uncover gene regulatory networks (GRNs) from experimental data. Primarily, reconstructing GRNs rely on gene expression data derived from gene microarrays.

As a major structure learning approach, Bayesian networks (BNs) describe a probabilistic graphical model by representing a set of random variables and conditional dependencies via a directed acyclic graph (DAG), which is widely used to analyze expression data\cite{Friedman:2000:UBN:332306.332355}\cite{NirFriedman02062004}. What's more, BNs provides a very flexible framework to fuse different types of data and prior knowledge together to derive a synthetic network in the process of GRNs inference\cite{RePEc:bpj:sagmbi:v:6:y:2007:i:1:n:15}.

BNs can cope with discrete or continuous expression levels, corresponding to underlying probabilistic model of multinomial distribution or multivariate Gaussian distribution. In general, discretization of continuous variables possesses advantages in computational efficiency, however, it would inevitably result in the loss of information\cite{celluarNetReview}. In comparison, BNs of continuous variables confront the challenge of computational complexity thus intractable and impractical to be applied into large-scale.

We have developed a Java framework, named \laga~ (\textbf{La}rge-Scale \textbf{G}ene \textbf{E}xpression), that provides a solution to reconstruct GRNs from continuous gene expression data for large scale of genes. The basic idea of our framework is motivated by the philosophy of divide-and-conquer\cite{LAMA}. Specifically, \laga~ recursively partitions genes into multiple overlapping communities with much smaller sizes, learns intra-community GRNs respectively before merge them altogether.

\section{Modules}

\laga~ contains four main functional modules: \\

$\bullet$ Partitioning large-scale network variables into multiple overlapping communities with much smaller sizes. We use Link Communities\cite{YY_LC_nature2010} for partition algorithm by utilizing the existing R package \emph{linkcomm}\cite{DBLP:journals/bioinformatics/KalinkaT11}.\\

$\bullet$ Sampling the community into multiple smaller sub-communities in the case the community size is still too large to perform practical Bayesian network learning. The sampling trategy borrows the idea from Random Node Neighbor (RNN)\cite{leskovec2006sampling}.\\

$\bullet$ Learning Bayesian network within each intra-community. We utilize existing R package \emph{deal} for Bayesian network learning with variables following conditional Gaussian Distribution\cite{deal}.\\

$\bullet$ Merging intra-community networks into a whole, by seeking an efficient merge order and resolving conflicts during mergence.

\begin{methods}
\section{Implementation}

\subsection{Partition Overlapping Gene Communities}
\label{partition}

\laga~ quantifies the correlation of pairwise genes directly from continuous gene expression values rather than discretization to avoid the loss of information. The correlation measurement is the absolute value of Pearson Coefficient. Thus all pairwise correlation values comprise a fully-connected network which is weighted and undirected.

In order to identify significant edges from the network, \laga~ prunes edges whose value is lower than certain truncate threshold $\mathcal {T}_{trunc}$, which is set to the mean value plus one standard deviation by default.

\laga~ partitions the weighted network into separate communities after the pruning. For the sake of convenience in mergence, communities are expected to be organized hierarchically. Meanwhile, for the sake of high coverage, communities are expected to maintain pervasive overlaps. \laga~ use Link Communities\cite{YY_LC_nature2010} for partition algorithm by utilizing the existing R package \emph{linkcomm}\cite{DBLP:journals/bioinformatics/KalinkaT11}.

\subsection{Sample Intra-Community Genes}
\label{sampling}

\laga~ tries to find out the candidate Markov Blanket set $MB(\mathcal {C})$ for each community $\mathcal {C}$. Conditioned on Markov Blanket, no other variables outside the community $\mathcal {C}$ could influence variables within the community. According to definition, the Markov Blanket of variable \emph{X} is composed of all parents of \emph{X}, all children of \emph{X} and all parents of \emph{X}'s children. In other words, Markov Blanket of \emph{X} should be closer to \emph{X} topologically than any other variables in networks. That is to say, topological closeness is related to the significance of edge weights.

For each variable \emph{X}, \laga~ selects for its Markov Blanket candidates by looking for adjacent nodes whose edge weights exceed certain threshold, which is the mean value plus one standard deviation by default.

\laga~ combines each community $\mathcal {C}$ with its Markov Blanket $MB(\mathcal {C})$ into a expanded community ${\mathcal {C}}'$. The size of ${\mathcal {C}}'$ would be much smaller than expectation, for intra-community variables are highly correlated due to partition. Chances are high that members of Markov Blanket is embodied in community $\mathcal {C}$ as well.

Finally, the sampling algorithm borrows the idea of Random Node Neighbor (RNN)\cite{leskovec2006sampling} by uniformly picking an unvisited variable as the starting node at random together with its neighbors, denoted by $\mathcal {S}$. The final sub-community for Bayesian network learning is $\mathcal {S} \cup MB(\mathcal {S})$, if the size of ultimate sub-community is still too large to learn, we keep removing neighbors within $\mathcal {S}$ until acceptable size.\\

\subsection{Learn Bayesian Network}
\label{learning}

For each sub-community, \laga~ learns the intra-community network by assuming the gene expression values are continuous variables following multivariate Gaussian Distribution. \laga~ utilizes existing R package \emph{deal} for Bayesian network learning with variables following conditional Gaussian Distribution\cite{deal}.

After learning sub-community networks, we integrate the networks into a uniform intra-community structure and resolve conflicts. We investigate the characteristics of error edges and find two major types of error:\\

$\bullet$ additional edges due to indirect interactions. This type of error is introduced by ARACNE\cite{DBLP:journals/bmcbi/MargolinNBWSFC06}.

$\bullet$ missing edges due to weak edge weight. This type of error originates from the adjacent nodes, maybe some nodes have relatively small PageRank value, in other words, they are periphery nodes; Or maybe some nodes are hubs.\\

To tackle these two major types of error, \laga~ first collect all candidate triplets of nodes. A candidate triplet contains three nodes, mutually-connected by the edge whose weight exceeds certain threshold value. Chances are high that these candidate triplets contain indirect interactions\cite{DBLP:journals/bmcbi/MargolinNBWSFC06}. After constructing a undirected, unweighted graph based on edges from these candidate triplets, \laga~ clusters this graph into sparsely-connected dense subgraphs by employing Link Communities\cite{YY_LC_nature2010} and re-learns the network for each cluster using the same approach.

\subsection{Ensemble Intra-Community Networks}
\label{merge}

\laga~ combines intra-community networks after learning individually from each community. The combination involves two concerns: (1) to find an efficient mergen order; (2) to resolve the conflicts during the merge.

Inspired by the idea of Huffman's Algorithm\cite{huf52}, \laga~ merges communities in a greedy strategy by constantly picking two communities with maximum Jaccard similarity coefficient\cite{jaccard}, denoted as $\mathcal {J}(\mathcal {C}_{i}, \mathcal {C}_{j})=\frac{\left | \mathcal {C}_{i}\bigcap \mathcal {C}_{j} \right |}{\left | \mathcal {C}_{i}\bigcup \mathcal {C}_{j}\right |}$.

The conflicts resolution follows the same strategy described in Section \ref{learning}.

\end{methods}

\section{Conclusion}

\laga~ is an scalable and parallel framework for the reconstruction of gene regulatory networks from continuous expression data. It provides an implementation in Java environment. \laga~ systematically divides all genes into multiple overlapping communities. Further, \laga~ employs sampling strategy to generate smaller sub-communities based on Markov Blanket candidates before performing Bayesian network learning. Finally, \laga~ merges intra-community GRNs together in a efficient order.

%
%

\bibliographystyle{plain}
\bibliography{citation}

\end{document}